\documentclass[letterpaper]{article} 
\usepackage[submission]{aaai24}  
\usepackage{times}  
\usepackage{helvet}  
\usepackage{courier}  
\usepackage[hyphens]{url}  
\usepackage{graphicx} 
\usepackage{natbib}  
\usepackage{caption} 
\usepackage{algorithm}
\usepackage{algorithmic}
\usepackage{amsmath,amsthm,amssymb,amsfonts}
\usepackage{booktabs}

\usepackage{newfloat}
\usepackage{listings}
\DeclareCaptionStyle{ruled}{labelfont=normalfont,labelsep=colon,strut=off} 
\floatstyle{ruled}
\newfloat{listing}{tb}{lst}{}
\floatname{listing}{Listing}
\title{VQ-NeRF: Vector Quantization Enhances Implicit Neural Representations }
\author {
    Yiying Yang\textsuperscript{\rm 1},
    Wen Liu\textsuperscript{\rm 2},
    Fukun Yin\textsuperscript{\rm 1,2},
    Xin Chen\textsuperscript{\rm 2},
    Gang Yu \textsuperscript{\rm 2}, 
    Jiayuan Fan \textsuperscript{\rm 1*}, 
    Tao Chen \textsuperscript{\rm 1}
}

\affiliations {
    Fudan University \textsuperscript{\rm 1}\\
    Tencent PCG \textsuperscript{\rm 2}
}

\usepackage{bibentry}

\usepackage{color}
\usepackage{xcolor}
\definecolor{fk}{HTML}{177cb0}

\definecolor{yyy}{HTML}{3CB371}

\begin{document}

\maketitle

\begin{abstract}
Recent advancements in implicit neural representations have contributed to high-fidelity surface reconstruction and photo-realistic novel view synthesis. However,  the computational complexity inherent in these methodologies presents a substantial impediment, constraining the attainable frame rates and resolutions in practical applications.
In response to this predicament, we propose VQ-NeRF, an effective and efficient pipeline for enhancing implicit neural representations via vector quantization. 
The essence of our method involves reducing the sampling space of NeRF to a lower resolution and subsequently reinstating it to the original size utilizing a pre-trained VAE decoder, thereby effectively mitigating the sampling time bottleneck encountered during rendering.
 Although the codebook furnishes representative features, reconstructing fine texture details of the scene remains challenging due to high compression rates. To overcome this constraint, we design an innovative multi-scale NeRF sampling scheme that concurrently optimizes the NeRF model at both compressed and original scales to enhance the network's ability to preserve fine details. Furthermore, we incorporate a semantic loss function to improve the geometric fidelity and semantic coherence of our 3D reconstructions.
Extensive experiments demonstrate the effectiveness of our model in achieving the optimal trade-off between rendering quality and efficiency ($cf.$ Figure~\ref{figure: rendering time comparisons.}). Evaluation on the DTU, BlendMVS, and H3DS datasets confirms the superior performance of our approach.


\end{abstract}

\section{Introduction}
\label{sec:Introduction}


\begin{figure}[t]
\centering
\includegraphics[width=1\columnwidth]{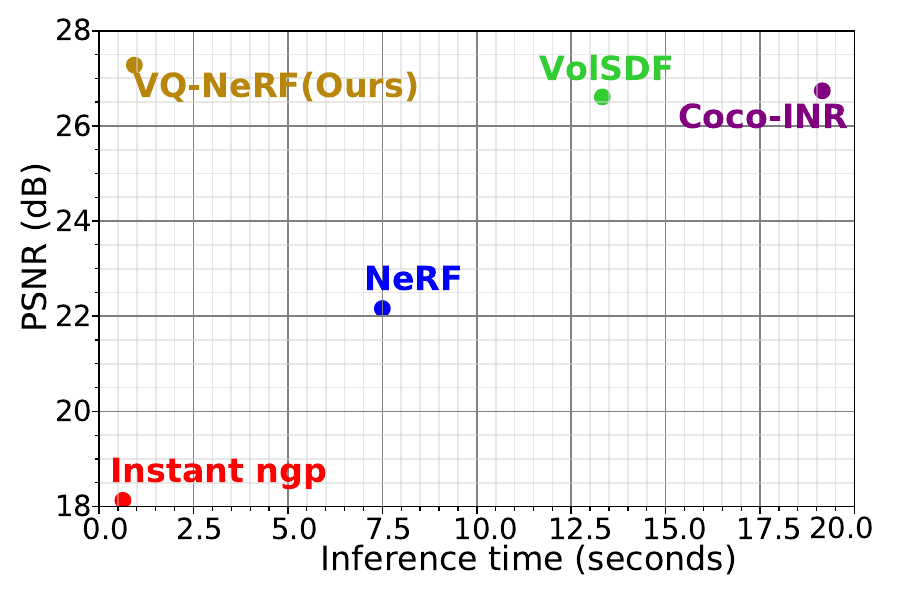} 
\caption{Rendering quality vs. inference time on the DTU
dataset. Our VQ-NeRF achieves the optimal trade-off between rendering quality and efficiency, compared with baselines NeRF~\cite{mildenhall2021nerf}, VolSDF~\cite{yariv2021volume}, Coco-INR~\cite{yin2022coordinates} and Instant ngp~\cite{muller2022instant}. }
\label{figure: rendering time comparisons.}
\end{figure}

\begin{figure*}[t]
\centering
\includegraphics[width=1\textwidth]{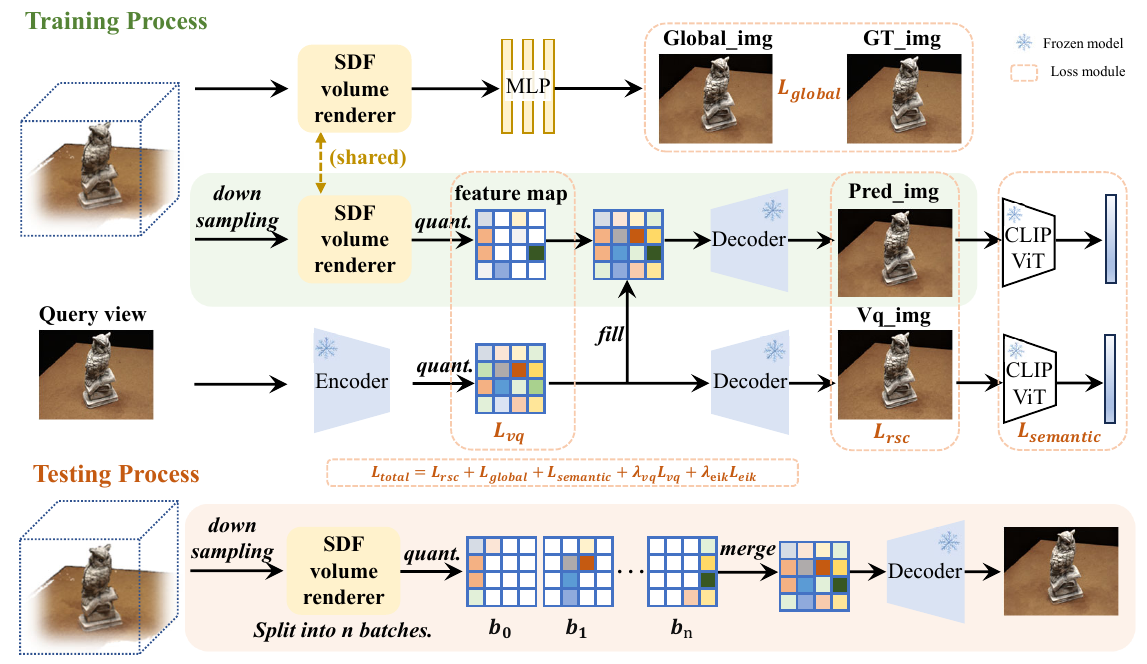} 
\caption{Overview of our VQ-NeRF. During the training process, we estimate and quantize feature encodings on a downsampled space,  ultimately decoding them into images of the original size with a pre-trained VAE decoder. Our multi-scale sampling scheme optimizes a parameter-shared SDF volume renderer and several additional MLP layers at the original scale to supplement the SDF volume renderer’s ability to represent texture details. Simultaneously, our method considers the semantic consistency between synthetic and real images, utilizing the CLIP model to enhance the realism of the scene. }
\label{figure: pipeline}
\end{figure*}

Implicit neural representations have demonstrated exceptional performance across a multitude of applications, including augmented reality, 3D modeling~\cite{xu2023wavenerf,wang2023learning}, and image synthesis~\cite{huang2023parametric,li2023one}. A series of methods, exemplified by NeRF~\cite{mildenhall2021nerf}, VolSDF~\cite{yariv2021volume}, and CoCo-INR~\cite{yin2022coordinates}, employ Multilayer Perceptrons (MLPs) and positional encoding to map coordinates to their corresponding color and density. However, these methods rely on sampling a vast number of points and processing them through MLPs during both the training and inference stages. This substantial computational burden presents a critical bottleneck, severely constraining the scope of practical applications.

To tackle this problem, an auxiliary explicit voxel grid has been utilized for the purpose of encoding local features, denoting a voxel-based approach. The voxel-based feature encoding has been implemented in various data structures, such as dense grids~\cite{sun2022direct}, octrees~\cite{liu2020neural,yu2021plenoctrees}, sparse voxel grids~\cite{fridovich2022plenoxels}, decomposed grids~\cite{chen2022tensorf}, and hash tables~\cite{muller2022instant}. For example, DVGO~\cite{sun2022direct} utilizes a learnable feature grid to reduce the size of the MLP network, thus decreasing the training time, and Instant-NGP~\cite{muller2022instant} employs multi-resolution hash encoding table for more memory and computationally-efficient approach. These representations succeed in efficiently reducing the time required for convergence and inference. However, explicit feature encoding methods still cannot faithfully represent the exact geometry, suffering from conspicuous noise and structural gaps. This can be attributed to the intrinsic ambiguity of the density-based volume rendering scheme.

In this paper, we aim to explore an effective and efficient pipeline for implicit neural representations that enhances both rendering quality and speed.
As a classic compression technique in signal processing, vector quantization (VQ),  a process that clusters multidimensional data into a finite set of representations, finds extensive applications in fields such as image processing~\cite{zhang2023lvqac}, image compression~\cite{feng2023nvtc}, etc. Previous approaches have attempted to combine VQ with implicit neural fields to achieve tasks such as super-resolution~\cite{or2022stylesdf}
However, these approaches struggle to expedite the rendering process of implicit neural fields. The essence of the problem lies in the fact that compressing the sampling space inevitably leads to a blurred scene representation and loss of image texture details. Relying solely on a pre-trained Variational Autoencoder (VAE) for upsampling can result in inconsistent perspectives.

To strike an optimal balance between rendering speed and quality, we propose VQ-NeRF: a novel framework that uniquely leverages Vector Quantization (VQ) to enhance the capabilities of Neural Radiance Fields (NeRF) in neural 3D surface representation (see Figure \ref{figure: pipeline}). To achieve this objective, (1) we pre-train a Variational Autoencoder (VAE), employing VQ to encode input data into a compact representation. Unlike existing NeRF methods, we do not sample the color and volume density of each pixel. Instead, we estimate and quantize feature encodings on a downsampled space, ultimately decoding them into images of the original size. (2) However, the decoder fails to recover the texture details of the scene, prompting the introduction of a multi-scale semantic consistency module. Specifically, we optimize a parameter-shared SDF volume renderer and several additional MLP layers at the original scale to supplement the SDF volume renderer's ability to represent texture details. Simultaneously, our method considers the semantic consistency between synthetic and real images, utilizing the CLIP model to enhance the realism of the scene.
By integrating these components, our VQ-NeRF framework aims to achieve an exceptional balance between rendering speed and quality, while generating realistic and semantically consistent scene representations and novel viewpoint synthesis results.

We conduct experiments on the DTU~\cite{jensen2014large}, BlendedMVS~\cite{yao2020blendedmvs} and H3DS~\cite{ramon2021h3d} datasets for quantitative and qualitative evaluations. Extensive experiments have shown the effectiveness of our framework in implicit scene representation and novel view synthesis. Our method outperforms the state-of-the-art approach in 3D surface reconstruction Coco-NeRF~\cite{yin2022coordinates} in terms of both rendering quality and significantly reduced rendering time (more than 10 times faster). We summarized our contributions as follows:

\begin{itemize}
    \item We propose a novel framework for implicit neural representation, leveraging a pre-trained VQ-VAE to compress the sampling space of implicit neural fields, thereby avoiding the computationally expensive per-pixel rendering process of traditional NeRF methods.
    \item We introduce a multi-scale semantic consistency module, composed of weight-shared global sampling and semantic consistency constraints, to address the loss of texture details caused by the compression of the sampling space and generate photo-realistic rendered images.
    \item Extensive experiments have demonstrated that our model achieves an optimal trade-off between rendering quality and speed compared to the latest methods.
\end{itemize}

\section{Related Work}
\label{sec: Related Work}

\textbf{Neural Volumetric Representations.} Neural volumetric representations are popular in 3D reconstruction and novel view synthesis. NeRF~\cite{mildenhall2021nerf} is based on the volume rendering equation and stores 3D information inside a neural network in the form of a compact Multi-layer Perceptron (MLP). However, the volume density estimated by NeRF does not enable high-quality surface extraction. 
Recently a family of methods instead focuses on neural surface representations and formulates compatible differentiable renderers. DVR~\cite{niemeyer2020differentiable} presents a differentiable renderer for implicit shape and texture representations, requiring only a multi-view RGB image and an object mask as supervision. IDR~\cite{yariv2020multiview} uses a pre-trained neuro reflector training end-to-end architecture, which can approximate surface reflected light and simulate various lighting conditions and materials by default. NeuS~\cite{wang2021neus} instead establishes a new method of training a bias-free neural SDF representation, while VolSDF~\cite{yariv2021volume} provides a novel parameterization for volume density, both contributing to more accurate surface reconstruction. UniSurf~\cite{oechsle2021unisurf} merges neural volume and surface rendering, enabling both within the same model. Coco-NeRF~\cite{yin2022coordinates} introduces a connection between each coordinate and the prior information, surpassing the previous MLPs-based implicit neural network. However, these methods depend on sampling an extensive number of points and subsequently processing them through MLP during both the training and inference stages, which often takes a long time (several hours) to optimize the network.

\noindent \textbf{Fast Neural Radiance Fields.} To tackle the problem of high computation costs, numerous methods for quicker convergence have been proposed. Mainly, DVGO~\cite{sun2022direct} utilizes a learnable feature grid to reduce the size of the MLP network, thus decreasing the training time, and Instant-NGP~\cite{muller2022instant} employs multi-resolution hash encoding table for more memory- and computationally-efficient approach. However, these explicit feature encoding methods, despite their improved efficiency, still face challenges in accurately representing the exact geometry of 3D objects or scenes, leading to conspicuous noise and structural gap. For reconstructing 3D surfaces, iMAP~\cite{sucar2021imap} and iSDF~\cite{ortiz2022isdf} have demonstrated that representing implicit surfaces through MLP could be done in real-time. Nevertheless, they depend on keyframe selection and active sampling, which sacrifice much of the details. Therefore, it's significant to devise an approach that can overcome these challenges, to achieve neural surface representations that are not only computationally efficient, but also accurately represent the geometry and semantic consistency of the 3D scenes.

\noindent \textbf{Vector Quantization.} 
As a classic compression technique in signal processing, vector quantization (VQ), a process that clusters multidimensional data into a finite set of representations, finds extensive applications in fields such as image processing~\cite{zhang2023lvqac}, image compression~\cite{feng2023nvtc}, etc. VQ-VAE~\cite{van2017neural} ﬁrst combines the VQ strategy with a variational autoencoder in generating images and speech. Then, VQGAN~\cite{esser2021taming} combines the codebook with adversarial learning to synthesize high-resolution images. 
However, these approaches struggle to expedite the rendering process of implicit neural fields. The essence of the problem lies in the fact that compressing the sampling space inevitably leads to a blurred scene representation and loss of image texture details. Our framework uniquely leverages VQ to enhance the capabilities of NeRF in neural 3D
surface representation, achieving an exceptional balance between rendering speed and quality. 

\section{Methodology}
\label{sec: Methodology}
In this section, we describe the overview of VQ-NeRF. Our framework can be seen in Figure~\ref{figure: pipeline}. In our framework, we pre-train a Variational Autoencoder (VAE), employing VQ to encode input data into a compact representation. Unlike existing NeRF methods, we do not sample the color and volume density of each pixel. Instead, we estimate and quantize feature encodings on a downsampled space, ultimately decoding them into images of the original size($c.f.$ Section~\ref{subsec:SDF-based}). However, the decoder fails to recover the texture details of the scene, prompting the introduction of a multi-scale semantic consistency module. Specifically, we optimize a parameter-shared SDF volume renderer and several additional MLP layers at the original scale to supplement the SDF volume renderer's ability to represent texture details ($c.f.$ Section~\ref{subsec: Multi-scale sampling scheme}). Simultaneously, our method considers the semantic consistency between synthetic and real images, utilizing the CLIP model to enhance the realism of the scene, and we will describe the optimization of our model in Section~\ref{subsec: Optimizatio}.


\subsection{Feature representation}
\label{subsec:SDF-based}
To tackle the challenge of computational burden, we reduce the sampling space of NeRF to a lower resolution and subsequently reinstate it to the original size utilizing a pre-trained VAE decoder. Specifically, we downsample the original image to one-fourth of the original size, and transmit the downsampled image to our SDF volume renderer.  Unlike existing NeRF methods, we estimate and quantize feature encodings on a downsampled space, ultimately decoding them into images of the original size. Our volume renderer takes a 3D query point $\mathbf{x}$, and a viewing direction $\mathbf{v}$ as input (sampled from downsampled space), and it outputs an SDF value $d(\mathbf{x})$, and feature vector $\mathbf{f}(\mathbf{x},\mathbf{v})$. The SDF value indicates the distance of the queried point from the surface boundary, with the sign serving as an indicator of the point's positioning either within or exterior to a watertight surface. A large positive value of the SDF would bias the sigmoid function towards zero, implying no density outside of the surface. Conversely, a high magnitude negative SDF value would push the sigmoid towards one, signifying maximal density interior to the surface. For each pixel point on the downsampled space, we query points on a ray that originates from the camera position (denoted as o) and follows the vector $\mathbf{r} = \mathbf{o} + t\mathbf{v}$, pointing in the direction of the camera, and calculate the feature map as follows:
\begin{equation}
\begin{aligned}
\mathbf{F}(\mathbf{r})=\int_{t_{n}}^{t_{f}} T(t) \sigma(\mathbf{r}(t)) \mathbf{f}(\mathbf{r}(t), \mathbf{v}) d t \\
\text{where} \quad T(t)=\exp \left(-\int_{t_{n}}^{t} \sigma(\mathbf{r}(s)) d s\right)
\end{aligned}
\end{equation}

\noindent The feature map generated from the SDF volume renderer is continuous in nature, capturing fine-grained geometric and topological information about the 3D scene. Inspired by vector quantization, we attempt to leverage vector quantization to extract key features from these continuous representations.  We derive the codebook for each dataset from training views by Vector-Quantized Variational AutoEncoder (VQVAE)~\cite{van2017neural}. The codebook contains critical information such as geometric features, topology, and texture attributes inherent in the data.  We denote our codebook as $\mathcal{E} = \{e_1, e_2, ..., e_{N} \} \in \mathbb{R}^{N \times n_q} $ , where $N$ is the number of prototype vectors,  $n_q$ is the dimension of each vector, and $e_i$ is each embedding vector. Given an image $x \in \mathbb{R}^{H \times W \times 3}$, VQ-VAEs learn a discrete codebook to represent observations as a collection of codebook entries $z_{q} \in \mathbb{R}^{h \times w \times n_q}$, where $h$ and $w$ are the height and width of the feature map generated from the output of SDF volume renderer, and $n_q$ is the dimensionality of quantized vectors in the codebook $\mathcal{E}$. Then a quantization $Q_g$ is performed onto its closest codebook entry $e_i$ for the continuous feature map $\hat{z}_{q}$ to obtain the discrete representation $z_{q}$ :
\begin{equation}
z_{q}= Q(\hat{z}_{q}) := \underset{e_{i} \in \mathcal{E}}{\operatorname{arg min}}\left\|\hat{z}_{q}-e_{i}\right\|_{2} 
\label{equ:vq-nerf}
\end{equation}
where $z_{q}$ contains the essential information needed for accurate 3D surface reconstruction, reducing computational overhead without sacrificing details. 
The model can be optimized by reducing the loss between the original image $I$ and the reconstructed image:
\begin{equation}
\mathbb{L} = \| x - \hat{x} \|^{2}  + 
\| sg(\hat{z}) - z_{q} \|^{2}_{2}  + \beta \| sg(z_{q}) - \hat{z} \|^{2}_{2}
\end{equation}
where sg denotes the stop-gradient operator, and $\beta$ is a hyperparameter for the third commitment loss. The first term is a reconstruction loss to estimate the error between the observed $x$ and the reconstructed $\hat{x}$. The second term is the codebook loss to optimize the entries in the codebook.

\subsection{Multi-scale Sampling Scheme}
\label{subsec: Multi-scale sampling scheme}
While the estimation and quantization on a downsampled space impressively decrease the rendering time, the pre-trained VAE decoder 
fails to recover the texture details of the scene. To overcome this challenge, we design a novel multi-scale NeRF sampling scheme that optimizes the NeRF model simultaneously at both compressed and original scales. Specifically, we optimize a parameter-shared SDF volume renderer and several additional MLP layers at the original scale (denoted as global sampling)to supplement the SDF volume renderer’s ability to represent texture details.  
Assuming a non-hollow surface, we convert the SDF value output from the SDF volume renderer into the 3D density fields $\sigma$,
\begin{equation}
 \sigma(\mathbf{x_g})=K_{\alpha}(d(x_g))=\frac{1}{\alpha} \cdot \operatorname{Sigmoid}\left(\frac{-d(\mathbf{x_g})}{\alpha}\right)
\end{equation}
where $\mathbf{x_g}$ represents a 3D query point sampled from the original image, and $\alpha$ is a learned parameter that serves to control the compactness of the density near the surface boundary. For global sampled on the original image, we query points on a ray that originates from the camera position (denoted as $o_g$) and follows the vector $\mathbf{r_g} = \mathbf{o_g} + t_g\mathbf{v_g}$, and calculate the RGB color as follows:

\begin{equation}
\begin{aligned}
\mathbf{C}(\mathbf{r_g})=\int_{t_{n}}^{t_{f}} T(t_g) \sigma(\mathbf{r_g}(t)) \mathbf{c_g}(\mathbf{r_g}(t), \mathbf{v_g}) d t \\
\text{where} \quad T(t_g)=\exp \left(-\int_{t_{n}}^{t} \sigma(\mathbf{r_g}(s)) d s\right)
\end{aligned}
\end{equation}

\begin{figure*}[t]
\centering
\includegraphics[width=1\textwidth]{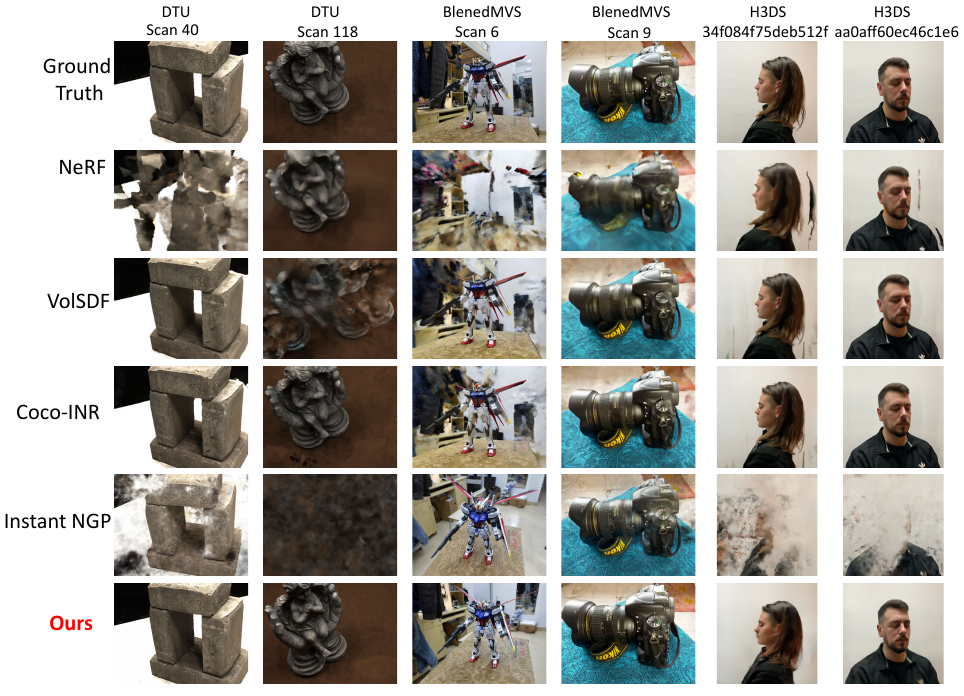} 
\caption{Qualitative comparisons on the DTU, BlendedMVS and H3DS dataset. Our VQ-NeRF consistently produces photo-realistic rendering results without any bad cases across different datasets. Compared to Coco-INR, our method can render a much clearer background, a feat that is challenging for networks based on coordinates and MLPs. Simultaneously, the quality of our rendering significantly surpasses that of Instant NGP, which primarily focuses on rapid rendering.}
\label{figure: Quality results}
\end{figure*}

\subsection{Optimization}
\label{subsec: Optimizatio}

\noindent \textbf{Semantic consistency loss.} 
Relying solely on a pre-trained Variational Autoencoder (VAE) for upsampling can result in inconsistent perspectives. To ensure semantic consistency, we consider the semantic
consistency between synthetic and real images, utilizing the
CLIP model to enhance the realism of the scene. The semantic consistency function, denoted as $L_{semantic}$, seeks to minimize the semantic discrepancy between the synthetic and real images. Leveraging the Contrastive Language-Image Pretraining (CLIP)~\cite{radford2021learning} model, our semantic consistency loss is calculated by comparing the high-dimensional semantic embeddings of the synthetic and real images, which can be obtained from the CLIP model. Specifically, the cosine distance between these two embeddings is used as the measure of semantic discrepancy:

\begin{equation}
L_{semantic}=1- cosine\left(E_{synthesic}-E_{real}\right)
\label{equation: semantic loss}
\end{equation}
where $E_{synthesic}$ represents the semantic embedding of the synthesized image and $E_{real}$ represents the semantic embedding of the real image. We refer to $L_{semantic}$ as a semantic consistency loss because it measures the similarity of high-level semantic features between synthesized and real views. 


\noindent \textbf{Multi-scale reconstruction loss.} With regard to our multi-scale sampling scheme, we denote our rendering loss on downsampled space as $L_{rsc}$ and rendering loss on the original scale as $L_{global}$ as shown in Figure~\ref{figure: pipeline}. 
The rendering loss enforces the rendered pixel color(denoted as $\hat{C_k}$) to be similar to the ground truth pixel color (denoted as $C_k$)), formulated as follows:

\begin{equation}
\begin{aligned}
L_{rsc}=\frac{1}{K} \sum_{k=1}^{K}\left\|\hat{C}_{k}-C_{k}\right\|_{2}   
\\
L_{global}=\frac{1}{K_g} \sum_{n=1}^{K_g}\left\|\hat{C}_{k_g}-C_{k_g}\right\|_{1}
\label{equation: reconstruction loss}
\end{aligned}
\end{equation}

\noindent \textbf{Vector quantization loss.}  The feature map from the downsampled space may lack some of the information of the original scale. To ensure the fidelity of our 3D reconstruction model, we supervise our model with a vector quantization loss function, aiming to minimize the discrepancies between the quantized feature map from the downsampled space and the traditional quantized representation from the pre-trained VAE encoder ( denoted as $L_{vq}$ in Figure~\ref{figure: pipeline}). Vector quantization loss is calculated as the mean of the absolute differences between two representations.

\begin{table*}[h]
  \centering
    \begin{tabular}{c|ccc|ccc|ccc}
    \hline
        & \multicolumn{3}{c|}{DTU} & \multicolumn{3}{c|}{BlendMVS} & \multicolumn{3}{c}{H3DS} \\
        & \textcolor[rgb]{ .2,  .2,  .2}{PSNR↑} & \textcolor[rgb]{ .2,  .2,  .2}{SSIM↑} & \textcolor[rgb]{ .2,  .2,  .2}{LPIPS↓} & \textcolor[rgb]{ .2,  .2,  .2}{PSNR↑} & \textcolor[rgb]{ .2,  .2,  .2}{SSIM↑} & \textcolor[rgb]{ .2,  .2,  .2}{LPIPS↓} & \textcolor[rgb]{ .2,  .2,  .2}{PSNR↑} & \textcolor[rgb]{ .2,  .2,  .2}{SSIM↑} & \textcolor[rgb]{ .2,  .2,  .2}{LPIPS↓} \\
    \hline
    \textcolor[rgb]{ .2,  .2,  .2}{NeRF~\cite{mildenhall2021nerf}} & \textcolor[rgb]{ .2,  .2,  .2}{22.162 } & \textcolor[rgb]{ .2,  .2,  .2}{0.790 } & \textcolor[rgb]{ .2,  .2,  .2}{0.306 } & \textcolor[rgb]{ .2,  .2,  .2}{16.523 } & \textcolor[rgb]{ .2,  .2,  .2}{0.667 } & \textcolor[rgb]{ .2,  .2,  .2}{0.272 } & \textcolor[rgb]{ .2,  .2,  .2}{21.185 } & \textcolor[rgb]{ .2,  .2,  .2}{0.861 } & \textcolor[rgb]{ .2,  .2,  .2}{0.144 } \\
    \textcolor[rgb]{ .2,  .2,  .2}{VolSDF~\cite{yariv2021volume}} & \textcolor[rgb]{ .2,  .2,  .2}{26.609 } & \textcolor[rgb]{ .2,  .2,  .2}{0.839 } & \textcolor[rgb]{ .2,  .2,  .2}{0.309 } & \textcolor[rgb]{ .2,  .2,  .2}{18.942 } & \textcolor[rgb]{ .2,  .2,  .2}{0.747 } & \textcolor[rgb]{ .2,  .2,  .2}{0.213 } & \textcolor[rgb]{ .2,  .2,  .2}{23.922 } & \textcolor[rgb]{ .2,  .2,  .2}{0.898 } & \textcolor[rgb]{ .2,  .2,  .2}{0.110 } \\
    \textcolor[rgb]{ .2,  .2,  .2}{Coco-NeRF~\cite{yin2022coordinates}} & \textcolor[rgb]{ .2,  .2,  .2}{26.738 } & \textcolor[rgb]{ .2,  .2,  .2}{0.852 } & \textcolor[rgb]{ .2,  .2,  .2}{0.298 } & \textcolor[rgb]{ .2,  .2,  .2}{19.594 } & \textcolor[rgb]{ .2,  .2,  .2}{0.764 } & \textcolor[rgb]{ .2,  .2,  .2}{0.201 } & \textcolor[rgb]{ .2,  .2,  .2}{\textbf{25.279} } & \textcolor[rgb]{ .2,  .2,  .2}{0.911 } & \textcolor[rgb]{ .2,  .2,  .2}{\textbf{0.098} } \\
    \textcolor[rgb]{ .2,  .2,  .2}{Instant ngp~\cite{muller2022instant}} & \textcolor[rgb]{ .2,  .2,  .2}{15.562 } & \textcolor[rgb]{ .2,  .2,  .2}{0.513 } & \textcolor[rgb]{ .2,  .2,  .2}{0.557 } & \textcolor[rgb]{ .2,  .2,  .2}{11.990 } & \textcolor[rgb]{ .2,  .2,  .2}{0.613 } & \textcolor[rgb]{ .2,  .2,  .2}{0.607 } & \textcolor[rgb]{ .2,  .2,  .2}{18.130 } & \textcolor[rgb]{ .2,  .2,  .2}{0.527 } & \textcolor[rgb]{ .2,  .2,  .2}{0.794 } \\
    \textcolor[rgb]{ .2,  .2,  .2}
    {\textbf{VQ-NeRF(Ours)}} & \textbf{27.278 } & \textbf{0.943 } & \textbf{0.191 } & \textcolor[rgb]{ .2,  .2,  .2}{\textbf{30.102 }} & \textcolor[rgb]{ .2,  .2,  .2}{\textbf{0.983 }} & \textcolor[rgb]{ .2,  .2,  .2}{\textbf{0.044 }} & 24.813  & \textbf{0.928 } & 0.226  \\
    \hline
    \end{tabular}%
    \caption{Quantitive comparison of our VQ-NeRF against baselines on the DTU, BlendedMVS and H3DS dataset. Our method outperforms existing approaches across various metrics on both the DTU and BlendMVS datasets. Additionally, on the H3DS dataset, our method is comparable to the top-performing Coco-INR approach. ↑ means the higher, the better, ↓ means the lower, the better.}
    \label{table: Quantitative results}%
\end{table*}

\noindent \textbf{Eikonal Loss.} This term guarantees the physical validity of the learned SDF~\cite{gropp2020implicit}: 
\begin{equation}
L_{e i k}=\frac{1}{N} \sum_{i=1}^{N}\left(\left\|\nabla f_{\phi}\left(x_{i}\right)\right\|-1\right)^{2}
\label{equation: eikonal loss}
\end{equation}

\noindent \textbf{Optimization.} We use the same loss functions as VolSDF~\cite{yariv2021volume}, along with our vector quantized loss, semantic consistency loss, and global reconstruction loss. Therefore, our total loss function is defined as:

\begin{equation}
L_{total}= L_{rsc} + L_{global} + L_{semantic} +\lambda_{vq} L_{vq}  + \lambda_{eik} L_{eik}  
\label{equation: total loss}
\end{equation}
where $\lambda_{vq}$= 1.0, $\lambda_{eik}$=0.1.

\section{Experiments}
\label{sec: Experiments}

\subsection{Experimental Settings}
\label{subsec: Experimental Settings}
\textbf{Datasets.} We conduct experiments on three different scene reconstruction datasets, which are known to be quite challenging and publicly available. These datasets include DTU~\cite{jensen2014large}, BlendMVS~\cite{yao2020blendedmvs} and H3DS~\cite{ramon2021h3d}. The DTU and BlendedMVS consist of datasets covering real objects presenting different properties in terms of material, appearance and geometric features. Every scene within the DTU dataset includes 49 to 64 posed images with a resolution of 1600 $\times$ 1200, and the BlendedMVS dataset contains 31 to 144 calibrated images, each with a resolution of 768 $\times$ 576. Consistent with previous studies~\cite{yariv2021volume,yin2022coordinates}, we conduct experiments analyzing 15 challenging scenes from the DTU dataset and 9 scenes from the BlendedMVS dataset. The H3DS dataset consists of 23 high-resolution, full-head 3D texture-scanned scenes with a variety of hairstyles operated under challenging lighting conditions. The dataset includes 64 calibrated images viewed from a 360-degree perspective.

\noindent \textbf{Baselines and metrics.} We compare our method against the recent state-of-the-art implicit neural 3D representation benchmarks, including NeRF~\cite{mildenhall2021nerf}, VolSDF~\cite{yariv2021volume}, CoCo-NeRF~\cite{yin2022coordinates} and Instant-NGP~\cite{muller2022instant}. For novel view synthesis, we adopt three common metrics: the Peak Signal Noise Ratio (PSNR), the Structural Similarity Index (SSIM)~\cite{wang2004image}, and the Learned Perceptual Image Patch Similarity (LPIPS)~\cite{zhang2018unreasonable}. For rendering time, we calculate the duration required to render an image across different benchmarks for fair comparisons.

\noindent \textbf{Implementation Details.} 
We utilize the pre-trained VQVAE~\cite{van2017neural} model for each dataset (DTU, BlendMVS, H3ds) with a codebook $\mathcal{E} \in \mathbb{R}^{2048 \times 16}$. We downsample the original image to a quarter of the original size. Specifically, the size of the original images on the H3DS dataset~\cite{ramon2021h3d} is 256$\times$256,  in our multi-scale sampling scheme, we downsample the input image size to a lower resolution of 64$\times$64, which greatly decreases the sampling space. The number of downsampled rays/pixels is 512 and the number of global sampled rays/pixels is 1024.
We adopt the same hierarchical sampling strategy as VolSDF~\cite{yariv2021volume}, including error constraints and geometric initialization. The Adam~\cite{kingma2014adam} variant of stochastic gradient descent was employed for parameter optimization, with the learning rate fixed at 0.0005. Our method is constructed within the Pytorch~\cite{paszke2019pytorch} framework. Each scene is trained on an Nvidia V100 GPU device for approximately 3-10 hours. 

\begin{figure*}[t]
\centering
\includegraphics[width=1\textwidth]{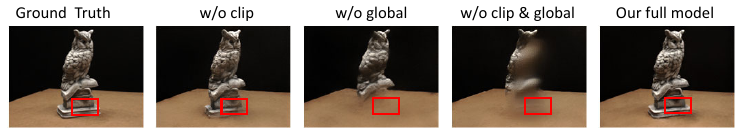} 
\caption{Qualitative results about the effectiveness of multi-scale sampling and semantic loss function on scene Scan122 from the DTU dataset.  }
\label{figure: Ablation experiments}
\end{figure*}

\subsection{Quantitative Comparisons.}
To comprehensively evaluate the performances, we compare our method with baselines on the DTU, BlenedMVS, and H3DS datasets. The quantitative results are presented in detail in Table~\ref{table: Quantitative results} and Table~\ref{tab: Rendering time of each baseline and our VQ-NeRF. }. Our method outperforms the other methods in terms of PSNR, SSIM, and LPIPS when only half of the views are accessible. We also report the rough inference time on the same hardware, $i.e.$, a single Nvidia V100 GPU. From Table~\ref{table: Quantitative results}, we can see that the inference time of Instant ngp is faster than other baselines, while the rendering qualities are unsatisfactory no matter in terms of PSNR, SSIM or LPIPS. In contrast, our VQ-NeRF not only produces the highest quality renderings on the DTU dataset and the BlendedMVS dataset but also can be rendered fast, $i.e.$ \textbf{0.935 seconds}, which is only inferior to Instant ngp. 

\begin{table}[htbp]
  \centering
   {\begin{tabular}{cccc}
    \toprule
    Method & Inference time(s)↓ \\
    \cmidrule(r){1-1} \cmidrule(r){2-2} 
    NeRF & 7.500 \\
    VolSDF  & 13.324  \\
    Coco-NeRF  & 19.147  \\
    Instant ngp &   0.543 \\
    VQ-NeRF (Ours) & 0.935  \\
    \hline
    \end{tabular}%
    }
    \caption{Quantitative results of the rough rendering time of each baseline and our VQ-NeRF. ↓ means the lower, the better. (The numbers may change with the GPU device. ) }
  \label{tab: Rendering time of each baseline and our VQ-NeRF. }%
\end{table}%

\subsection{Qualitative Comparisons.}
We further qualitatively compared our VQ-NeRF with baselines in Figure~\ref{figure: Quality results}. As illustrated in Figure~\ref{figure: Quality results}, we can see that the results of Instant ngp suffer from blurriness for structure and texture details, which is the tough challenge for explicit neural representations to represent neural surfaces. 
Methods such as NeRF and VolSDF, which rely solely on coordinates and MLPs, exhibit poor robustness and encounter failure cases in certain scenarios.
Despite the incorporation of additional scene priors on each coordinate in Coco-INR, the limited expressive power of MLP networks results in producing blurry rendering images for background regions.
Meanwhile, Compared with NeRF and VolSDF, Coco-INR improve the rendering quality while significantly increasing the inference time, $i.e.$, and 19.147 seconds per view. In contrast, our VQ-NeRF decreases the inference time by 14$\times$ compared with VolSDF and by 20$\times$ compared with Coco-INR, maintaining the rendering quality.


\subsection{Ablation Studies and Analysis}

\textbf{Effectiveness of multi-scale sampling and semantic loss function.} To verify the effectiveness of our multi-scale sampling scheme, as is demonstrated in Section~\ref{subsec: Multi-scale sampling scheme}, we conduct ablation studies on different setups: removing the semantic loss function, denoted as w/o clip; conducting experiments only on the downsampling scale, denoted as w/o global; conducting experiments on the downsampling scale without semantic loss function, denoted as w/o clip $\&$ global. The comparison results are shown in Table~\ref{tab: Effectiveness of multi-scale sampling and semantic loss function} and Figure~\ref{figure: Ablation experiments}, which implies that our multi-scale sampling scheme significantly enhances the network’s ability to preserve fine details. The ablation experiments on the semantic loss function imply the semantic coherence of our 3D reconstructions by means of the decrease of the LPIPS. As shown in Figure~\ref{figure: Ablation experiments}, the picture in the third column fails to recover the texture details of the scene, which demonstrates the significance of our multi-scale sampling scheme.  Simultaneously, the picture in the second column exhibits a notable loss in semantic consistency.

\begin{table}[htbp]
  \centering
   {\begin{tabular}{cccc}
    \toprule
    Method & PSNR↑ & SSIM↑ & LPIPS↓ \\
    \cmidrule(r){1-1} \cmidrule(r){2-4} 
    w/o clip & 28.525 & 0.933 & 0.220 \\
    w/o global  & 25.212   & 0.819   & 0.319 \\
    w/o clip $\&$ global & 22.480 & 0.685 & 0.458 \\
    \cmidrule(r){1-1} \cmidrule(r){2-4}
    \textbf{ Ours} & \textbf{32.056} & \textbf{0.972} & \textbf{0.143} \\
    \hline
    \end{tabular}%
    }
    \caption{Experiment results about the effectiveness of multi-scale sampling and semantic loss function on the DTU dataset on the scene "scan122". ↑ means the higher, the better, ↓ means the lower, the better.}
  \label{tab: Effectiveness of multi-scale sampling and semantic loss function}%
\end{table}%

\begin{table}[htbp]
  \centering
   {\begin{tabular}{cccc}
    \toprule
    Method & PSNR↑ & SSIM↑ & LPIPS↓ \\
    \cmidrule(r){1-1} \cmidrule(r){2-4} 
    4096$\times$16  & 32.026 & 0.968 & \textbf{0.143} \\
    1024$\times$16  & 30.002 & 0.926 & 0.171 \\
    
    2048$\times$32  & 31.211   & 0.929   & 0.155 \\
    2048$\times$8  & 31.742 & 0.964 & 0.162 \\
    \cmidrule(r){1-1} \cmidrule(r){2-4}
    \textbf{ Ours(2048$\times$16)} & \textbf{32.056} & \textbf{0.972} & \textbf{0.143} \\
    \hline
    \end{tabular}%
    }
    \caption{Experiment results about the effectiveness of the codebook size on the DTU dataset on the scene "scan122". ↑ means the higher, the better, ↓ means the lower, the better.} 
  \label{tab: Impact of the codebook size.}%
\end{table}%

\noindent \textbf{Impact of the codebook size.}  We report the results about the different codebook sizes during the quantization of our model in Table~\ref{tab: Impact of the codebook size.}. Experiments show that our method VQ-NeRF presents the best performance with the codebook size 2048$\times$16.
When the codebook size is too small, there may not be enough prototypes to represent the scene features adequately. Conversely, when the codebook is too large, the VAE may overfit to the training viewpoints and lack generalization across different viewpoints.

\section{Conclusion and limitations}

In this paper, we propose the VQ-NeRF that utilizes vector quantization to accelerate implicit neural rendering, achieving an optimal trade-off between speed and quality. The essence of our approach lies in reducing the sampling time during rendering by compressing the sampling space of the implicit neural field and leveraging VQ-VAE to obtain images at their original size. To counteract the feature loss caused by spatial compression, we design a multi-scale sampling technique and employ semantic consistency evaluation to enhance the representation of details and realism in the synthesized images. Extensive experiments validate that our VQ-NeRF outperforms previous methods in synthesizing photo-realistic novel viewpoints and achieving better quantitative evaluations.

 Despite the accelerated rendering process achieved by our VQ-NeRF, it still follows the conventional approach of scene-specific optimization, which requires a significant amount of training time. In future work, we aim to explore solutions that can establish a general representation for each scene, enabling better generalization across different scenes. This would help reduce the training time and improve the efficiency of our method.

\bibliography{aaai24}

\end{document}